%% file: main.tex
\documentclass[lettersize,journal]{IEEEtran}

\usepackage{cite}
\usepackage{amsmath,amsfonts}
\usepackage{algorithmic}
\usepackage{algorithm}
\usepackage{array}
\usepackage[caption=false,font=normalsize,labelfont=sf,textfont=sf]{subfig}
\usepackage{textcomp}
\usepackage{stfloats}
\usepackage{url}
\usepackage{verbatim}

\usepackage[colorlinks,
            linkcolor=blue,
            anchorcolor=blue,
            citecolor=blue]{hyperref}

\usepackage{graphicx}
\usepackage{amssymb}
\usepackage{booktabs}
\usepackage{array}
\usepackage{color}
\usepackage{colortbl} 
\usepackage{svg}
\usepackage{float}
\usepackage{subcaption}
\usepackage{multirow}
\usepackage{amssymb} 
\usepackage{makecell}
\usepackage{bm}
\usepackage{float}

\usepackage{marvosym}

\newcommand{\etal}{ \emph{et al.} }
\newcommand{\eg}{\emph{e.g.}, }
\newcommand{\ie}{ \emph{i.e.}, }

\begin{document}

\title{RepVideo: Rethinking Cross-Layer Representation for Video Generation}

\author{Chenyang~Si$^\dag$,
        Weichen~Fan$^\dag$,
        Zhengyao~Lv,
        Ziqi~Huang,
        Yu Qiao,
        and~Ziwei~Liu~\textsuperscript{\Letter} 

\thanks{Chenyang Si is with the S-Lab, Nanyang Technological University, Singapore, 639798. \protect\\ E-mail: chenyang.si@ntu.edu.sg.}
\thanks{Weichen Fan is with the S-Lab, Nanyang Technological University, Singapore, 639798. \protect\\ E-mail: fanweichen2383@gmail.com.}
\thanks{Zhengyao Lv is with Shanghai Artificial Intelligence Laboratory, China. \protect\\ E-mail: cszy98@gmail.com.}
\thanks{Ziqi Huang is with the S-Lab, Nanyang Technological University, Singapore, 639798. \protect\\ E-mail:  ziqi002@ntu.edu.sg.}
\thanks{Yu Qiao is with Shanghai Artificial Intelligence Laboratory, China. \protect\\ E-mail:  qiaoyu@pjlab.org.cn.}
\thanks{Ziwei Liu is with the S-Lab, Nanyang Technological University, Singapore, 639798. \protect\\ E-mail: ziwei.liu@ntu.edu.sg.}
\thanks{ $\dag$ Equal contribution.}

}



\maketitle

\begin{abstract}
    Video generation has achieved remarkable progress with the introduction of diffusion models, which have significantly improved the quality of generated videos. However, recent research has primarily focused on scaling up model training, while offering limited insights into the direct impact of representations on the video generation process. In this paper, we initially investigate the characteristics of features in intermediate layers, finding substantial variations in attention maps across different layers. These variations lead to unstable semantic representations and contribute to cumulative differences between features, which ultimately reduce the similarity between adjacent frames and negatively affect temporal coherence.
    To address this, we propose RepVideo, an enhanced representation framework for text-to-video diffusion models. By accumulating features from neighboring layers to form enriched representations, this approach captures more stable semantic information. These enhanced representations are then used as inputs to the attention mechanism, thereby improving semantic expressiveness while ensuring feature consistency across adjacent frames. Extensive experiments demonstrate that our RepVideo not only significantly enhances the ability to generate accurate spatial appearances, such as capturing complex spatial relationships between multiple objects, but also improves temporal consistency in video generation. Project page: \url{https://vchitect.github.io/RepVid-Webpage}.
\end{abstract}

\begin{IEEEkeywords}
Video Generation, Diffusion Models, Video Diffusion Models, Transformer.
\end{IEEEkeywords}


\input{sec/1_introduction}

\input{sec/2_related_work}

\input{sec/3_method}
\input{sec/4_experiments}

\input{sec/5_conclusion}

\ifCLASSOPTIONcompsoc
  \section*{Acknowledgments}
\else
  \section*{Acknowledgment}
\fi


This study is supported by the Ministry of Education, Singapore, under its MOE AcRF Tier 2 (MOET2EP20221- 0012), NTU NAP, and under the RIE2020 Industry Alignment Fund – Industry Collaboration Projects (IAF-ICP) Funding Initiative, as well as cash and in-kind contribution from the industry partner(s).


\ifCLASSOPTIONcaptionsoff
  \newpage
\fi



%
{ \small
  \bibliographystyle{IEEEtran}
  \bibliography{egbib}
}




%

\begin{IEEEbiography}[{\includegraphics[width=1in,height=1.25in,clip,keepaspectratio]{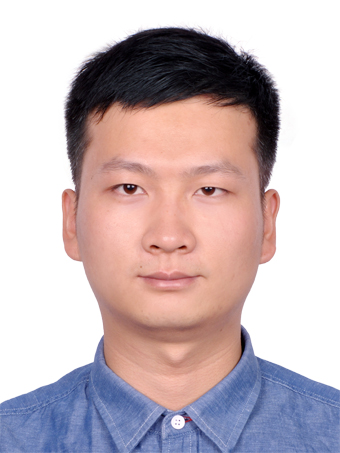}}]{Chenyang Si}
received the B.S. degree from Zhengzhou University, Zhengzhou, China, in 2016, and the Ph.D. degree from the National Laboratory of Pattern Recognition (NLPR), Institute of Automation, Chinese Academy of Sciences (CASIA), Beijing, China, in 2021. Currently, he is a research fellow at Nanyang Technological University (NTU) Singapore. His research lies at the intersection of deep learning and computer vision, including vision-based human perception (pose and action), few-shot learning, self-supervised learning, semi-supervised learning, network architecture, and image/video generation.
\end{IEEEbiography}

\begin{IEEEbiography}[{\includegraphics[width=1in,height=1.25in,clip,keepaspectratio]{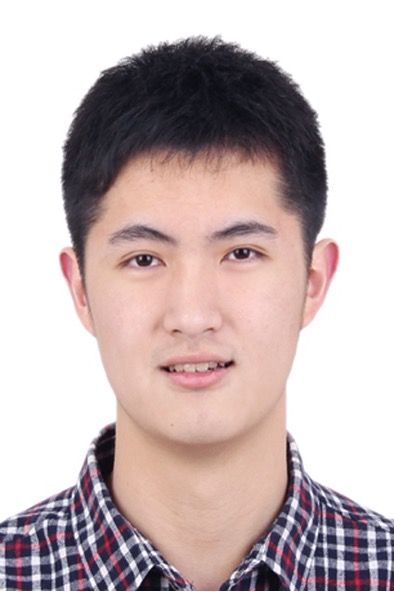}}]{Weichen Fan}
received the bachelor's degree from University of Electronic Science and Technology of China (UESTC), and the master's degree from National University of Singapore (NUS). He is currently working toward the Ph.D. degree with MMLab@NTU, Nanyang Technological University, supervised by Prof. Ziwei Liu.
His research interests include image/video generation, robotics, and robotic simulation.
\end{IEEEbiography}

\begin{IEEEbiography}
[{\includegraphics[width=1in,height=1.25in,clip,keepaspectratio]{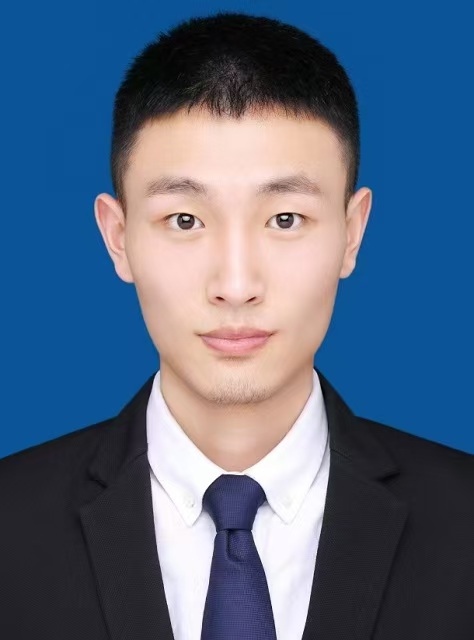}}]{Zhengyao Lv}  is currently a Ph.D. student at the University of Hong Kong, supervised by Prof. Kenneth K.Y. Wong. He received his Master's degree in 2022 and his Bachelor's degree in 2020 from Harbin Institute of Technology. His current research interests focus on visual generation.
\end{IEEEbiography}

\begin{IEEEbiography}
[{\includegraphics[width=1in,height=1.25in,clip,keepaspectratio]{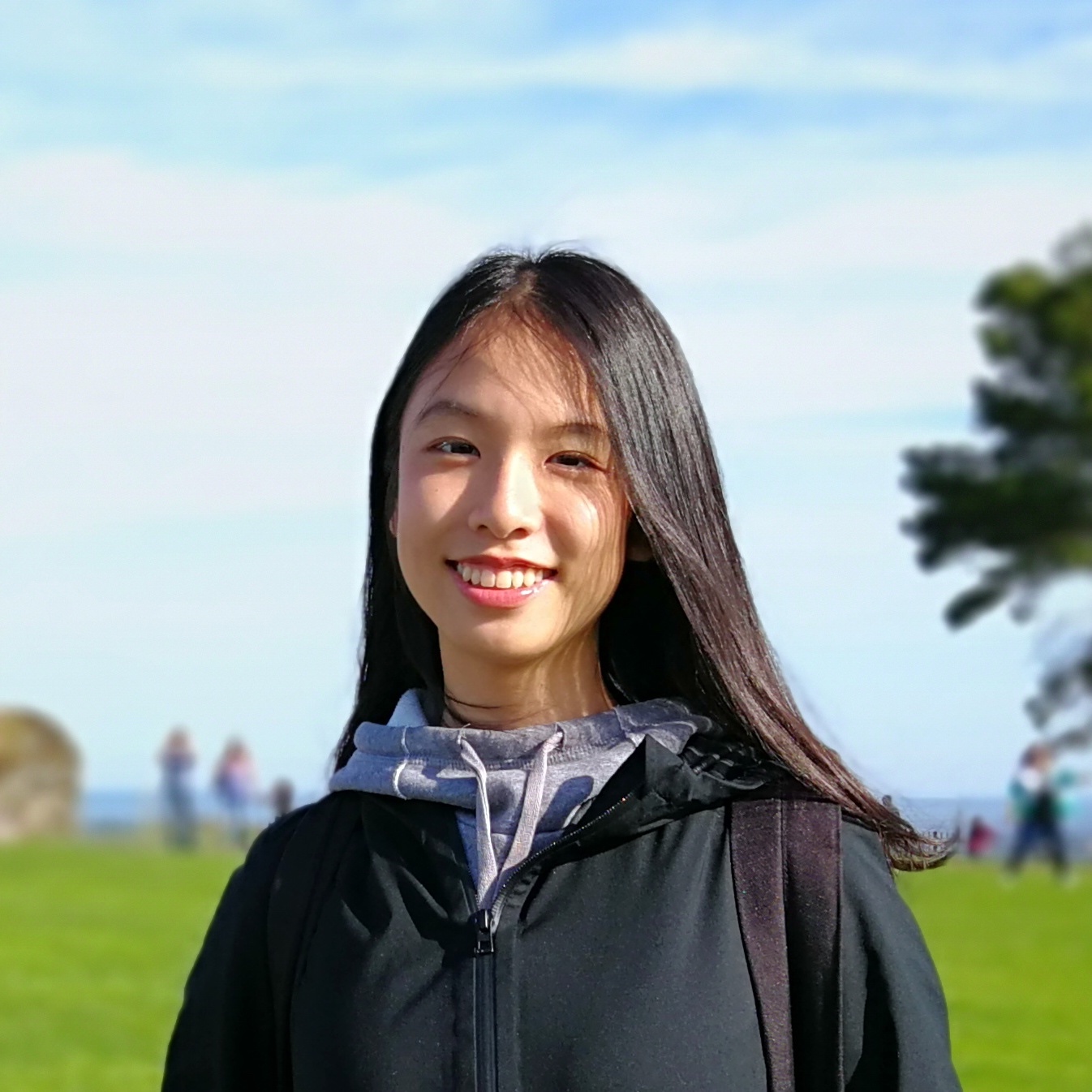}}]{Ziqi Huang} is currently a Ph.D. student at MMLab@NTU, Nanyang Technological University (NTU), supervised by Prof. Ziwei Liu. She received her Bachelor's degree from NTU in 2022. Her current research interests include visual generation and evaluation. She is awarded Google PhD Fellowship 2023.
\end{IEEEbiography}

\begin{IEEEbiography}
[{\includegraphics[width=1in,height=1.25in,clip,keepaspectratio]{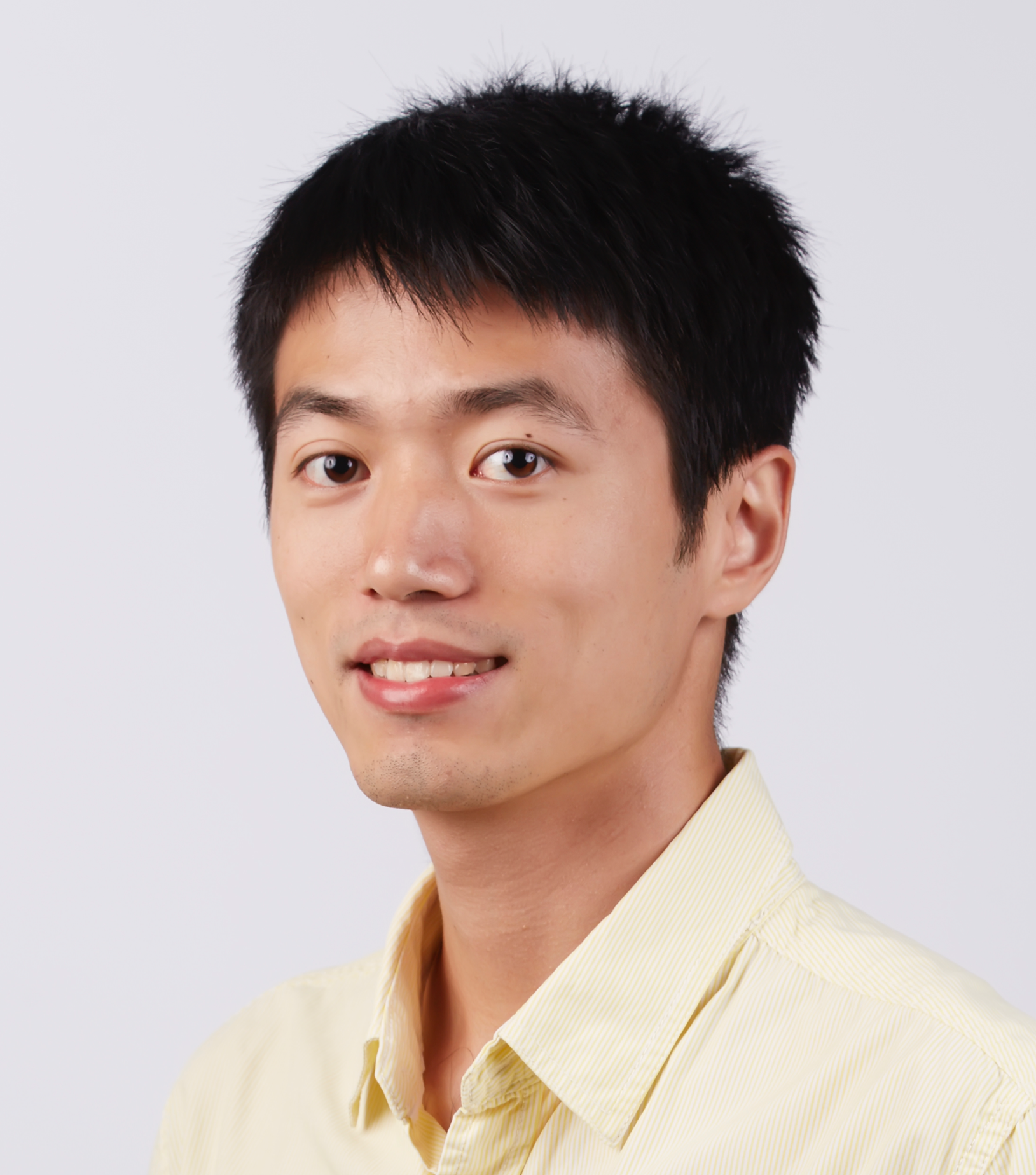}}]{Ziwei Liu} is currently a Nanyang Assistant Professor at Nanyang Technological University, Singapore. His research revolves around computer vision, machine learning and computer graphics. He has published extensively on top-tier conferences and journals in relevant fields, including CVPR, ICCV, ECCV, NeurIPS, ICLR, ICML, TPAMI, TOG and Nature - Machine Intelligence. He is the recipient of Microsoft Young Fellowship, Hong Kong PhD Fellowship, ICCV Young Researcher Award, HKSTP Best Paper Award and WAIC Yunfan Award. He serves as an Area Chair of CVPR, ICCV, NeurIPS and ICLR, as well as an Associate Editor of IJCV.
\end{IEEEbiography}






\end{document}

%% file: sec/1_introduction.tex

\section{Introduction}
\label{sec:introduction}

\IEEEPARstart{R}{ecently}, the research community has witnessed remarkable advancements in generative models, particularly in the domain of text-to-image generation (T2I)~\cite{esser2021taming,ramesh2021zero,ding2021cogview,lin2021m6,yu2022scaling}. These breakthroughs have garnered significant attention, showcasing the potential of generative models in producing diverse and realistic images from textual descriptions. Notably, diffusion models~\cite{ho2020denoising, song2020denoising,rombach2022high,peebles2023scalable,chen2023pixart,podellsdxl} have emerged as a significant development, enabling the generation of high-fidelity images through the iterative diffusion of information within the model. Inspired by these achievements, researchers are now exploring the application of diffusion models in the realm of text-to-video (T2V) generation~\cite{singer2022make,ho2022imagen,an2023latent,zhou2022magicvideo,blattmann2023align,wang2023lavie,blattmann2023stable,zhang2024show,ma2024latte,guo2023animatediff,harvey2022flexible,chen2023videocrafter1,wang2023dreamvideo,menapace2024snap,jiang2024videobooth}, aiming to generate visually compelling and contextually coherent videos with textual descriptions.

In the realm of video generation, producing coherent and high-quality videos is inherently complex and demanding. Unlike static images, videos require the generation of sequential frames that exhibit spatial fidelity and temporal continuity. However, the availability of high-quality video datasets is significantly limited compared to image datasets, making it challenging to obtain diverse and representative training data. Additionally, generating a single frame in a video involves not only capturing visual details but also preserving smooth transitions and consistent motion across frames. Consequently, video generation entails substantial computational requirements due to the increased number of frames and the need for effective modeling of temporal dependencies. 

To overcome these challenges and advance the field of video generation, early video diffusion methods sought to extend state-of-the-art T2I models to text-to-video (T2V) generations, such as Make-A-Video~\cite{singer2022make}, Imagen Video~\cite{ho2022imagen}, Latent-Shift~\cite{an2023latent}, MagicVideo~\cite{zhou2022magicvideo}, Video LDM~\cite{blattmann2023align}, and Lavie~\cite{wang2023lavie}. These methods primarily adapt image diffusion models by incorporating temporal modules such as temporal attention~\cite{he2022latent}, spatio-temporal attention~\cite{wu2022tune}, directed temporal attention~\cite{zhou2022magicvideo}, temporal shift~\cite{an2023latent}, and 3D CNNs~\cite{blattmann2023align}, and fine-tune them on RGB video data.

\begin{figure*}[t]
	\centering
	\includegraphics[width=0.99\textwidth]{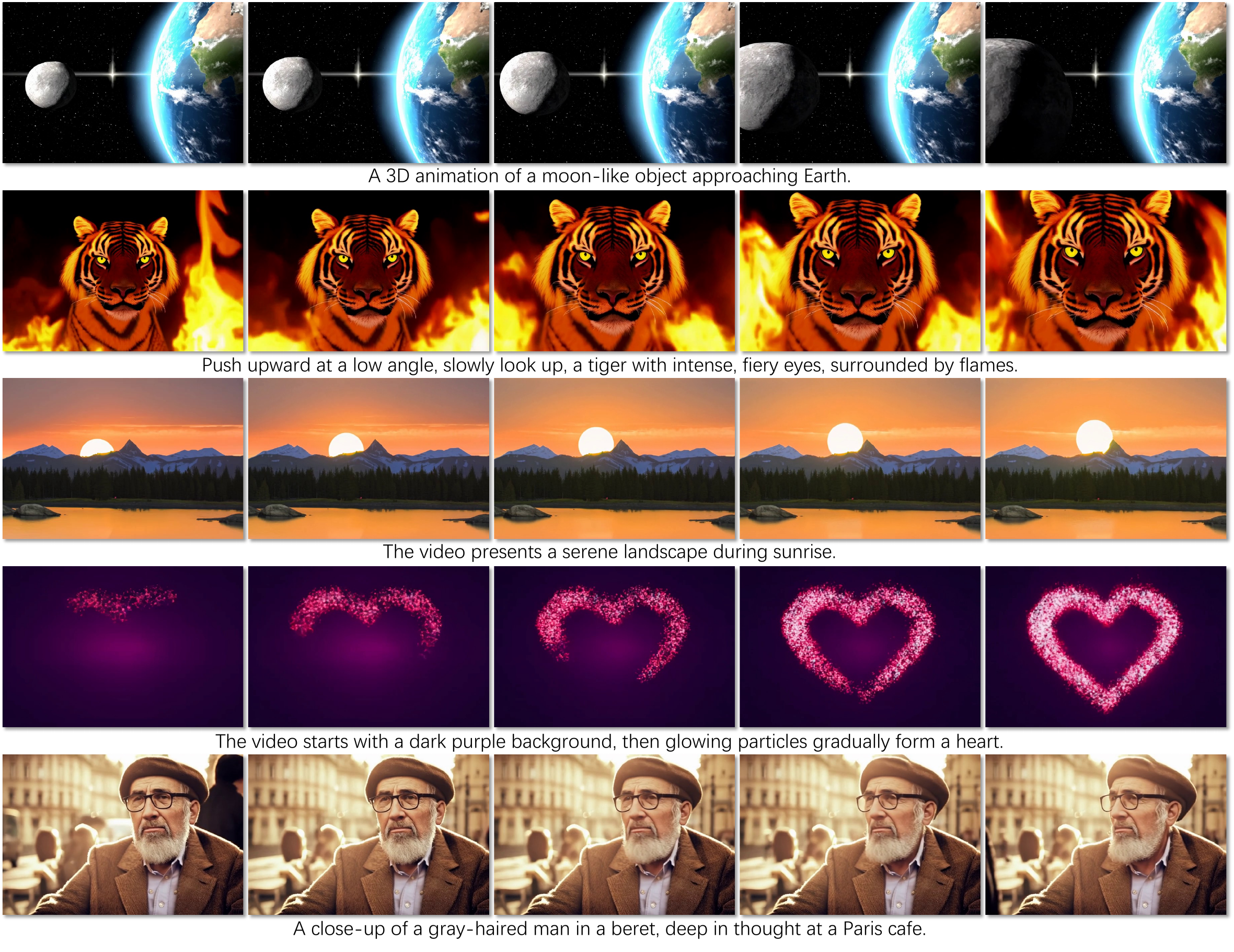}
	\caption{\textbf{The examples generated by RepVideo.} RepVideo effectively generates diverse videos with enhanced temporal coherence and fine-grained spatial details.}
	\label{fig_teaser}
\end{figure*}

With the recent success of Diffusion Transformers (DiTs)~\cite{peebles2023scalable,chen2023pixart,esser2024scaling} in text-to-image generation, which leverage transformers as the backbone of diffusion models, exploration has extended to text-to-video generation. A significant advancement is achieved with the OpenAI Sora~\cite{videoworldsimulators2024}, which significantly scales a spatial-temporal transformer model, marking a milestone in video generation technology. Subsequently, transformer-based approaches, such as CogVideoX~\cite{yang2024cogvideox}, Vidu~\cite{bao2024vidu} and MovieGen~\cite{polyak2024movie}, have further advanced video generation. These methods typically use a 3D VAE to compress input videos along spatial and temporal dimensions, yielding a video latent representation. This latent is then flattened into the token sequence. A transformer network captures spatial and temporal information within this sequence, effectively generating complex videos. Despite these significant successes, most studies have primarily focused on scaling up model training, \ie model size, dataset scale, and compute budget, while offering limited insights into how representations directly impact the video generation process.

In this paper, we conduct an in-depth analysis of transformer representations within video diffusion models, focusing on their impact on spatial appearance and temporal consistency in video generation. Achieving both fine spatial detail and smooth temporal transitions is essential for producing high-quality videos. To this end, we first investigate the spatial expressiveness of attention maps across transformer layers. Our findings reveal that the attention maps in each transformer layer exhibit substantial differences, with each layer focusing on distinct token information. While this enables the model to capture diverse spatial features, the lack of coordination across layers leads to fragmented feature representations, weakening the model’s ability to form coherent spatial semantics within individual frames.

Building on this, we further extend our investigation to examine how these attention mechanisms affect temporal consistency. We analyze the evolving similarity between adjacent frame features across transformer layers, as this similarity serves as a key indicator of temporal coherence. Our findings show that as layer depth increases, the similarity between adjacent frame features gradually decreases. This phenomenon is attributed to the progressive feature differentiation introduced by attention mechanisms, which accumulates across layers. As a result, the diminished frame-to-frame similarity disrupts temporal coherence, leading to artifacts such as motion discontinuities or blur in the generated videos, ultimately impairing their overall quality.

Based on the above insights, we propose a novel architecture, termed RepVideo, to enhance the video representations for text-to-video diffusion models. 
The attention module of each transformer layer captures distinct local information, leading to a diverse set of intermediate feature representations. Hence, we explore leveraging these rich features to enhance the semantic consistency and quality of generated videos. Specifically, We introduce the feature cache module that aggregates the features from multiple adjacent transformer layers. We apply a mean aggregation strategy across all collected features to achieve a stable semantic representation. 
Finally, this aggregated representation is combined with the original transformer input through a gating mechanism, producing an enhanced feature input for each transformer layer. With the guidance of the stable semantic representation, RepVideo can maintain consistent feature details across frames, alleviating inconsistencies between adjacent frames. This leads to enhanced temporal coherence, while also improving the model's capacity for generating details, thereby resulting in higher quality and more visually consistent video generations, as shown in Fig.~\ref{fig_teaser}. 

We conducted extensive experiments to evaluate the effectiveness of RepVideo. The experimental results demonstrate that RepVideo significantly enhances both temporal coherence and spatial detail generation, achieving competitive performance in qualitative and quantitative metrics. 

Our contributions are summarized as follows:
\begin{itemize}
    \item We investigate the transformer representations in video diffusion models, revealing that substantial variations in attention maps across layers lead to fragmented spatial semantics and reduced temporal consistency, which negatively impact video quality.

    \item We propose RepVideo, a framework that leverages a feature cache module and a gating mechanism to aggregate and stabilize intermediate representations, enhancing both spatial detail and temporal coherence.

    \item Extensive experiments demonstrate that RepVideo achieves competitive performance in both temporal consistency and spatial quality, validating its effectiveness for video generation.
    
\end{itemize}

%% file: sec/2_related_work.tex
\section{Related work}
\subsection{Generative Models}

Generative models have been extensively studied and have achieved significant advancements. Early research focused on the application of Generative Adversarial Networks (GANs), which excelled at generating visually realistic and semantically coherent images. For example, AttnGAN~\cite{xu2018attngan} introduced an attention mechanism to focus on specific image regions, enhancing fine-grained details. DM-GAN~\cite{zhu2019dm} incorporated a dynamic memory module to refine image synthesis quality, while DF-GAN~\cite{tao2022df} proposed a simple yet effective framework for one-stage high-resolution image generation. On the other hand, a different line of research for text-to-image generation has emerged with the framework of Vector Quantized Variational AutoEncoders (VQ-VAE). These methods tokenize images into discrete visual representations, forming the foundation for models such as DALL-E~\cite{ramesh2021zero}, CogView~\cite{ding2021cogview}, M6~\cite{lin2021m6}, and Parti~\cite{yu2022scaling}. These models leverage VQ-VAE~\cite{van2017neural} or VQGAN~\cite{esser2021taming} to encode images and employ large-scale language models to map textual descriptions to visual tokens, enabling the generation of semantically aligned, high-quality images. More recently, Denoising Diffusion Probabilistic Models (DDPM)~\cite{ho2020denoising} have garnered significant attention for their ability to produce high-fidelity images with robust text-to-image alignment~\cite{ song2020denoising,rombach2022high,peebles2023scalable,chen2023pixart,podellsdxl, ramesh2022hierarchical}. Notable examples include GLIDE~\cite{nichol2022glide}, which adopts a classifier-free guidance strategy for improved generation, and Imagen~\cite{saharia2022photorealistic}, which introduces an efficient UNet architecture for diffusion models. 
table Diffusion~\cite{rombach2022high, podellsdxl} further optimizes computational efficiency by applying diffusion in the latent space while maintaining high image quality. Building on these advancements, transformer-based diffusion models~\cite{peebles2023scalable,chen2023pixart,esser2024scaling} have emerged as a powerful alternative to the UNet architecture, achieving superior performance in generating high-resolution and semantically consistent images.

\subsection{Video Generative models}

While significant advancements have been made in image generation, video generation from textual descriptions presents additional challenges. Unlike static images, videos require not only capturing spatial details but also modeling temporal dynamics and maintaining coherent transitions across frames. Similar to the development of image generation, early text-to-video (T2V) methods primarily employed Generative Adversarial Networks (GANs), such as those proposed in \cite{li2018video, mittal2017sync, pan2017create}. These early approaches focused on constrained domains, such as sequences of moving digits or specific human actions. However, generating videos in more complex and diverse domains introduces substantial challenges, as it requires modeling intricate spatial-temporal relationships and handling greater variability in visual content. 
As an alternative approach, auto-regressive methods have also been exploited for video generation. GODIVA~\cite{wu2021godiva} introduces a three-dimensional sparse attention mechanism into VQ-VAE for an open-domain text-to-video generation. VideoGPT~\cite{yan2021videogpt} adapts VQ-VAE and Transformer models to generate realistic samples from complex natural video datasets. TATS as proposed by Ge\etal~\cite{ge2022long}, extends the 2D-VQGAN to a 3D-VQGAN for modeling videos and employs a hierarchical approach with a frame interpolation transformer to generate long, coherent, and high-quality videos. The paper by Wu\etal~\cite{wu2022nuwa} introduces NUWA, a multimodal model that leverages auto-regressive architectures to facilitate the generation of videos and images from textual inputs. CogVideo \cite{hong2022cogvideo}, an autoregressive transformer model, leverages the pre-trained text-to-image generative model to the text-to-video generation. Phenaki~\cite{villegas2022phenaki} is capable of generating variable-length videos conditioned on a sequence of open domain text prompts.
Diffusion models, a recent class of models, offer a promising framework for generation tasks, including T2V generation which has been the focus of recent research efforts~\cite{ho2022video,wang2023lavie,ho2022imagen,zhou2022magicvideo,wu2022tune,an2023latent,luo2023videofusion}. Ho\etal\cite{ho2022video} propose a video diffusion model by using factorized space-time attention blocks for joint image-video training. 
Imagen Video~\cite{ho2022imagen} extends the T2I diffusion model of Imagen~\cite{saharia2022photorealistic} to generate high-fidelity videos through a cascade of video diffusion models. MagicVideo~\cite{zhou2022magicvideo} introduces the directional attention and the adaptor module into Stable Diffusion~\cite{rombach2022high} for generating videos.  Make-A-Video~\cite{singer2022make} extends the pre-trained T2I diffusion model through a spatiotemporally factorized diffusion model. Tune-A-Video~\cite{wu2022tune} finetunes the pre-trained text-to-image diffusion model on a single video for video synthesis. Video LDM~\cite{blattmann2023align} leverages pre-trained image modes and turns them into video generators by inserting temporal layers that learn to align images in a temporally consistent manner.Latent-Shift~\cite{an2023latent} propose a temporal shift module to leverage a T2I model as-is for T2V generation without adding any new parameters. LVDM~\cite{he2022latent} uses a hierarchical latent video diffusion model to generate longer videos beyond the temporal training length. VideoFusion~\cite{luo2023videofusion} presents a decomposed diffusion process that separates per-frame noise into shared base noise and residual noise, improving the generation of high-quality videos. Most recently, transformer-based diffusion models have marked a significant milestone in video generation. Models like Sora~\cite{videoworldsimulators2024} have demonstrated remarkable performance, showcasing the potential of transformer diffusion architectures. Additionally, models such as CogVideoX~\cite{yang2024cogvideox}, and MovieGen~\cite{polyak2024movie}, have achieved impressive results, pushing the boundaries of T2V generation with enhanced spatial-temporal coherence and scalability. Despite showing some promising results, these methods provide limited insights into how intermediate representations directly influence the video generation process.
In this paper, we conduct a comprehensive analysis of transformer representations within video diffusion models, focusing on their role in video quality and temporal coherence.

%% file: sec/3_method.tex
\begin{figure*}[!h]
	\centering
	\includegraphics[width=0.9\textwidth]{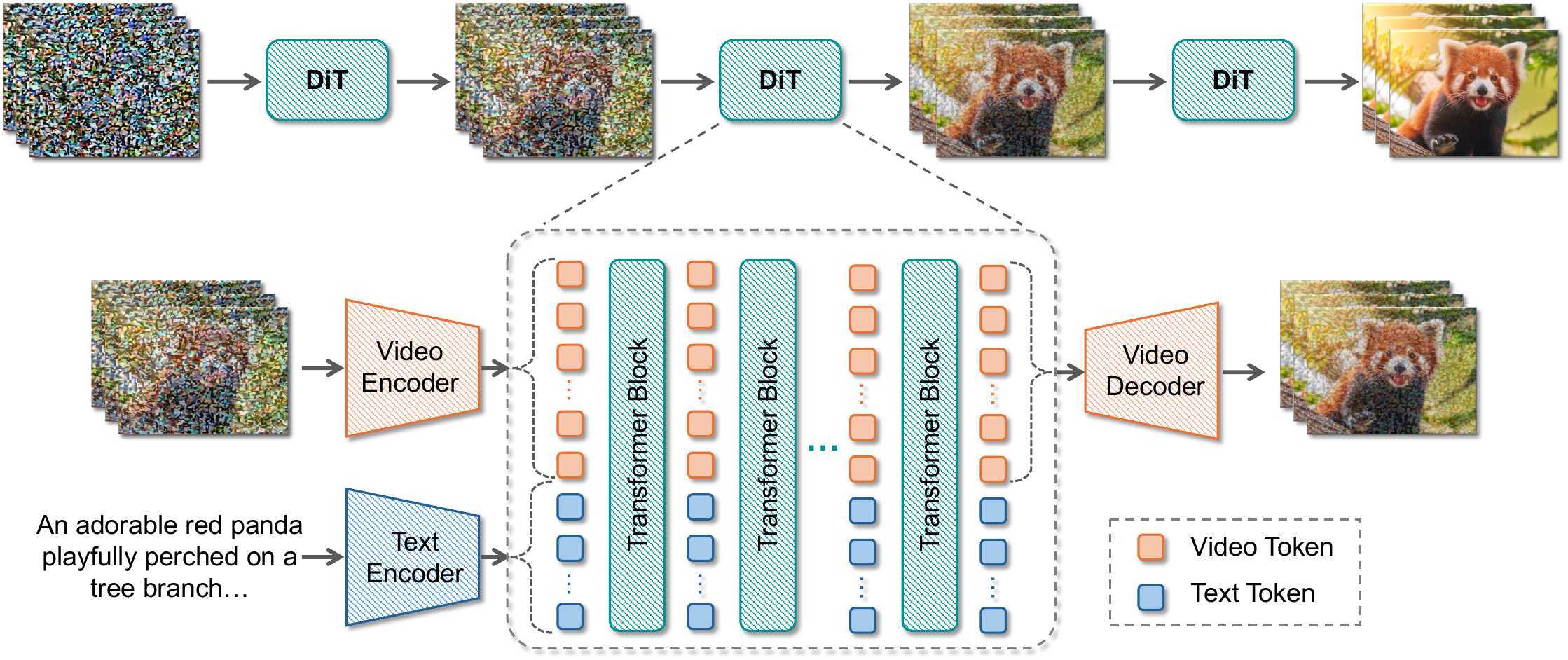}
	\caption{\textbf{The architecture of recent transformer-based video diffusion models}. These methods typically consist of three core components: a 3D VAE, the text encoder, and a transformer network. }
	\label{fig_DIT_pipline}
\end{figure*}

\section{Methodology}

\subsection{Preliminaries}

\subsubsection{Denoising Diffusion Probabilistic Models}

Denoising Diffusion Probabilistic Models (DDPM), as introduced in \cite{ho2020denoising}, encompass a diffusion process and a denoising process for data modeling. The diffusion process in DDPM is represented by a sequence of $T$ steps. At each step $t$, Gaussian noise is progressively injected into the data distribution $\boldsymbol{x}_0 \sim q(\boldsymbol{x}_0)$ through a Markov chain according to a variance schedule $\beta_1,...,\beta_T$:
\begin{align}
    q(\boldsymbol{x}_{1:T}|\boldsymbol{x}_{0}) = \prod_{t=1}^T q(\boldsymbol{x}_t|\boldsymbol{x}_{t-1}) 
\end{align}
\begin{align}
    q(\boldsymbol{x}_t|\boldsymbol{x}_{t-1}) = \mathcal{N}(\boldsymbol{x}_t;\sqrt{1-\beta_t}\boldsymbol{x}_{t-1},\beta_t \mathcal{I}) 
\end{align}
where the variances $\beta_t$ are kept constant throughout the diffusion process, as suggested by \cite{ho2020denoising}. The diffusion process consists of $T$ steps, where random noise is progressively added to each sample $\boldsymbol{x}_{t-1}$ to generate the next sample $\boldsymbol{x}_t$. The magnitude of the added noise at each step is determined by the predefined variance schedule $\beta_t$.

The denoising process reverses the above diffusion process to the underlying clean data $\boldsymbol{x}_{t-1}$ given the noisy input $\boldsymbol{x}_t$:
\begin{align}
    p_\theta(\boldsymbol{x}_{0:T}) = p(\boldsymbol{x}_{T}) \prod_{t=1}^T p_\theta(\boldsymbol{x}_{t-1}|\boldsymbol{x}_{t})
\end{align}
\begin{align}
    p_\theta(\boldsymbol{x}_{t-1}|\boldsymbol{x}_{t}) = \mathcal{N}(\boldsymbol{x}_{t-1};\boldsymbol{\mu}_\theta(\boldsymbol{x}_{t}, t), \boldsymbol{\Sigma}_\theta(\boldsymbol{x}_{t}, t)) 
\end{align}
The $\boldsymbol{\mu}_\theta$ and $\boldsymbol{\Sigma}_\theta$ are estimated using a denoising model $\epsilon_\theta$. This denoising model is commonly implemented as a time-conditional UNet architecture, which incorporates residual blocks and self-attention layers. The denoising model $\epsilon_\theta$ is trained to remove noise and enhance the fidelity of the generated samples. During training, the objective is to maximize a simplified version of the variational bound:
\begin{align}
    \label{eqn_diffusion_loss}
    \mathcal{L} = \mathbb{E}_{\boldsymbol{x}, \epsilon\sim\mathcal{N}(0,1), t} \lbrack \Vert \epsilon - \epsilon_\theta(\boldsymbol{x}_t, t) \Vert_2^2	\rbrack
\end{align}
where $\epsilon$ denotes the unscaled noise.

\subsubsection{Latent Diffusion Models}

Latent Diffusion Models \cite{rombach2022high} (LDMs) are a type of diffusion model that models the distribution of the latent space of images. The training process of latent image diffusion models consists of two stages. 

In the first stage, an autoencoder is employed to learn a compressed and meaningful representation of input images. The autoencoder is composed of an encoder $\mathcal{E}$ and a decoder $\mathcal{D}$. The encoder takes an input image $\boldsymbol{x} \in \mathbb{R}^{H \times W \times 3}$ and maps it to a lower-dimensional latent space, producing a latent representation $\boldsymbol{z} = \mathcal{E}(\boldsymbol{x}) \in \mathbb{R}^{h \times w \times c}$. Here, $H$ and $W$ represent the height and width of the input image, while $h$, $w$, and $c$ represent the dimensions of the feature maps in the latent space. The decoder then reconstructs the original image from this latent representation, generating a reconstructed image $\boldsymbol{\Tilde{x}} = \mathcal{D}(\boldsymbol{z}) \in \mathbb{R}^{H \times W \times 3}$. In the LDMs, two commonly used architectures for the autoencoder component are VQGAN~\cite{esser2021taming} and VAE~\cite{kingma2013auto}. In the latent space, the diffusion model of the second stage can be trained with the objective in Eq.~\ref{eqn_diffusion_loss}, where the latent representation $\boldsymbol{z}$ replaces the original image $x$, \ie $\mathbb{E}_{\boldsymbol{z}, \epsilon\sim\mathcal{N}(0,1), t} \lbrack \Vert \epsilon - \epsilon_\theta(\boldsymbol{z}_t, t) \Vert_2^2	\rbrack$. Compared to the high-dimensional pixel space, LDMs offer advantages in terms of parameter count and memory consumption, while maintaining similar performance. This property makes LDMs a more efficient and practical choice for generative tasks.

\begin{figure*}[t]
	\centering
	\includegraphics[width=0.99\textwidth]{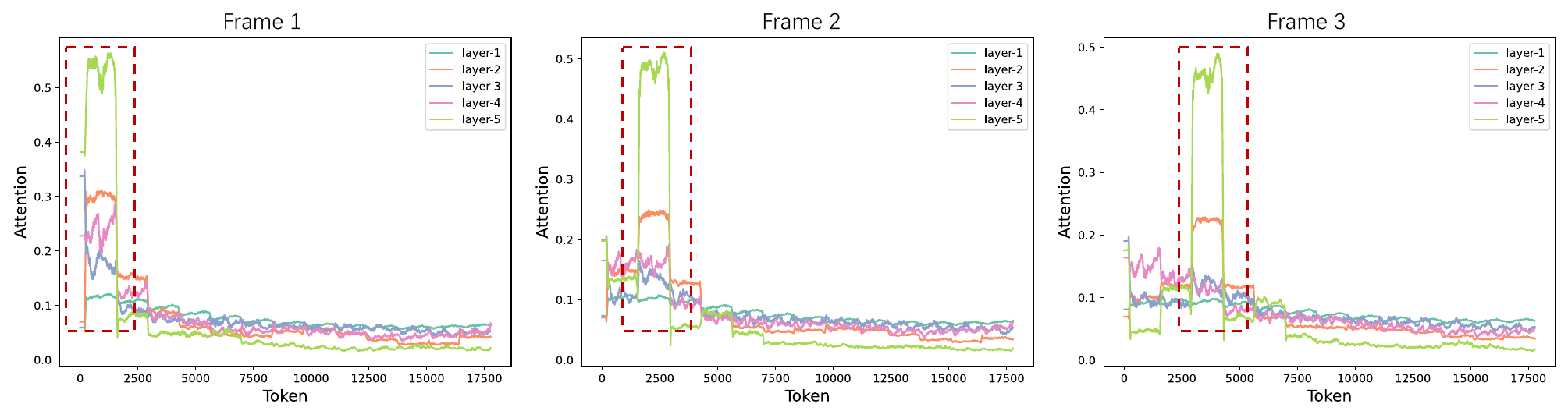}
	\caption{\textbf{The visualization of the attention distribution of each frame’s token across the entire token sequence.} The results highlight significant variations in attention distributions across layers, with deeper layers focusing more on tokens from the same frame and exhibiting weaker attention to tokens from other frames. }
	\label{fig_attention_distribution}
\end{figure*}

\begin{figure*}[h]
	\centering
	\includegraphics[width=0.99\textwidth]{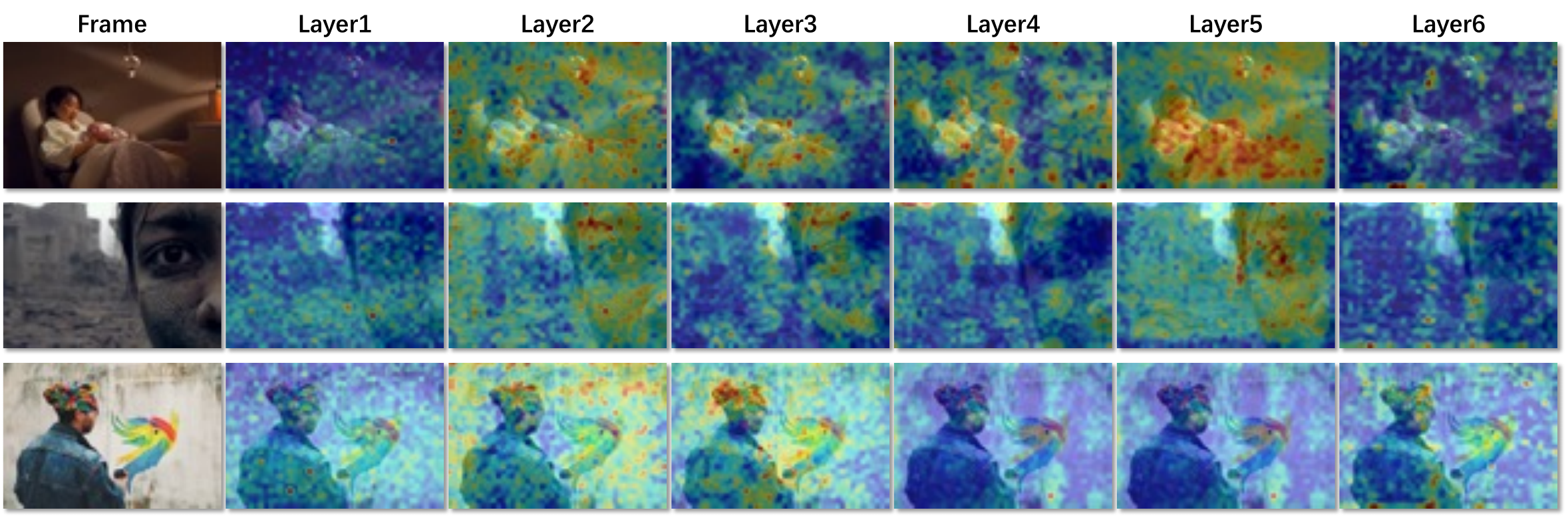}
	\caption{\textbf{The visualization of attention maps across transformer layers.}  Each layer attends to distinct regions, capturing diverse spatial features. However, the lack of coordination across layers results in fragmented feature representations, weakening the model’s ability to establish coherent spatial semantics within individual frames.}
	\label{fig_attention_maps_cog}
\end{figure*}

\subsection{Rethinking the Representation of Video Diffusion Models}
Recently, diffusion transformers have demonstrated significant success in text-to-video generation, such as CogVideoX~\cite{yang2024cogvideox}, and MovieGen~\cite{polyak2024movie}. These methods typically consist of three core components: a 3D VAE, the text encoder, and a transformer network, as shown in Fig~\ref{fig_DIT_pipline}. The 3D VAE is used to compress the video data along spatial and temporal dimensions, yielding a compact latent representation that allows efficient processing of higher video resolutions and larger numbers of frames, while significantly reducing GPU memory usage. The text encoder processes the input text prompt, converting it into a set of embeddings that capture the semantic meaning and guide the entire video generation process. The video latent representation is then flattened into a sequence of tokens, which, along with the text embedding tokens, are fed into the transformer network. By leveraging the transformer’s powerful attention mechanism, it can learn complex spatial and temporal relationships within the video sequence, ensuring that the generated frames are coherent, consistent, and aligned with the semantic information provided by the text prompt. By integrating these components, diffusion transformer-based models have shown remarkable improvements in generating high-resolution, long-duration videos that are both temporally consistent and semantically aligned with the input prompt. Despite these advancements, most studies have primarily focused on scaling up model training, such as increasing model size, training dataset scale, and computational resources, while providing limited insights into how intermediate representations directly influence the video generation process. This emphasis on scaling overlooks the potential benefits of understanding and optimizing the internal feature representations, which could lead to more efficient and coherent video generation without merely relying on increased model capacity.


\begin{figure*}[t]
	\centering
	\includegraphics[width=0.99\textwidth]{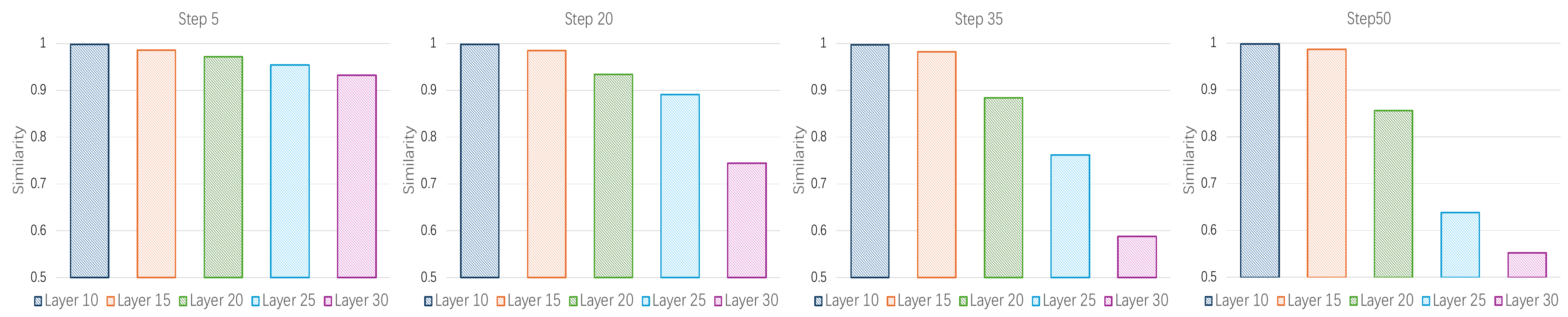}
	\caption{\textbf{The average similarity between adjacent frame features across diffusion layers and denoising steps.}  The similarity decreases as layer depth increases for a given denoising step, indicating greater differentiation in deeper layers. Additionally, similarity between adjacent frames declines as the denoising process progresses.}
	\label{fig_similarity_layer_step}
\end{figure*}

To gain a comprehensive understanding of the role that intermediate representations play in video generation, we perform an in-depth analysis of transformer representations within video diffusion models, particularly focusing on their impact on spatial expressiveness and temporal consistency.
We begin by analyzing the spatial expressiveness of attention maps across transformer layers. Recent models, as illustrated in Fig~\ref{fig_DIT_pipline}, integrate both video latent tokens and text embedding tokens into a unified token sequence, which is then processed by the transformer using a full attention mechanism to capture relationships among these tokens. To understand which regions in the attention maps contribute to semantic meaning, we visualize the attention distribution of each frame's token across the entire token sequence, as shown in Fig~\ref{fig_attention_distribution}. The results reveal significant variations in attention distributions across layers, with each layer focusing on distinct regions and learning different aspects of the features. Additionally, we find that as layer depth increases, the attention corresponding to each frame's token becomes more concentrated on the tokens from the same frame, with relatively weaker attention to tokens from other frames.

This observation suggests that analyzing the attention maps of each frame's token alone can effectively represent the overall attention characteristics of the sequence. Thus, we can focus on the self-attention maps of individual frame tokens to capture the key global attention features, simplifying the analysis while still providing meaningful insights into how attention is distributed across different spatial and temporal regions in the video. As shown in Fig~\ref{fig_attention_maps_cog}, we visualize the attention maps for a single frame token across different layers. The visualization reveals that the attention maps from different layers exhibit substantial differences, with each layer attending to distinct regions and capturing different feature information. As the network depth increases, these attention mechanisms, also lead to progressive differentiation of features. While this enables the model to capture diverse spatial features, the lack of coordination across layers leads to fragmented feature representations, weakening the model’s ability to form coherent spatial semantics within individual frames.

Building on this, we extend our investigation to examine how these attention mechanisms affect temporal consistency. We analyze the evolving similarity between adjacent frame features across transformer layers, as this similarity serves as a critical indicator of temporal coherence. Fig~\ref{fig_similarity_layer_step} visualizes the average similarity between adjacent frame features across different diffusion layers at various denoising steps. The analysis reveals two key observations. First, for a given denoising step, the similarity between adjacent frame features decreases as layer depth increases. This suggests that deeper layers introduce progressively diverse features, leading to greater differentiation between frame features. Second, when comparing across denoising steps, the similarity between adjacent frames decreases as the denoising process progresses. For example, the average similarity is higher at earlier steps (\eg Step 5) but gradually decreases at later steps (\eg Step 5). This trend indicates that, while the denoising process enriches the video features with more diverse semantic content, it also increases variability between adjacent frames, reducing temporal coherence and potentially introducing motion artifacts in the generated videos.


\begin{figure*}[h]
	\centering
	\includegraphics[width=0.9\textwidth]{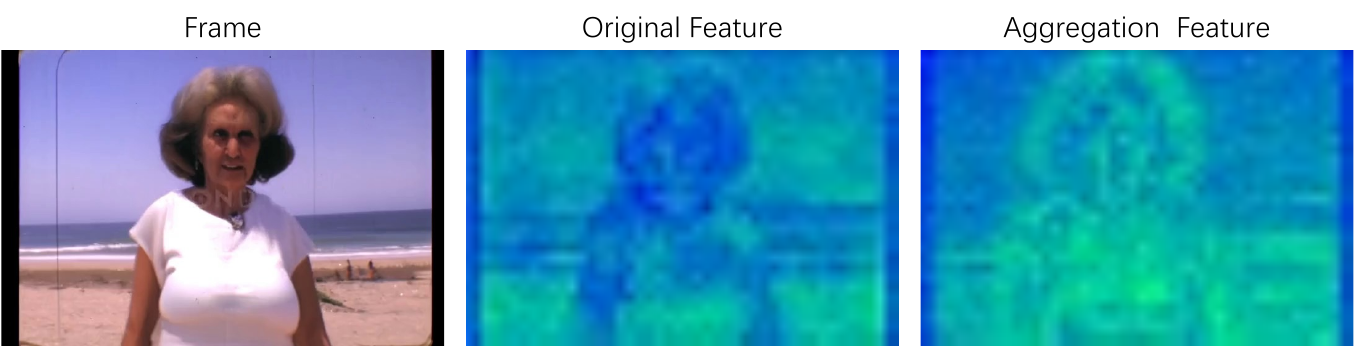}
	\caption{\textbf{The comparison of the original feature maps from a standard transformer layer with those obtained after aggregation in the Feature Cache Module.} The aggregated features demonstrate more comprehensive semantic information and clearer structural details.}
	\label{fig_original_mean_feat}
\end{figure*}

\begin{figure*}[h]
	\centering
	\includegraphics[width=0.99\textwidth]{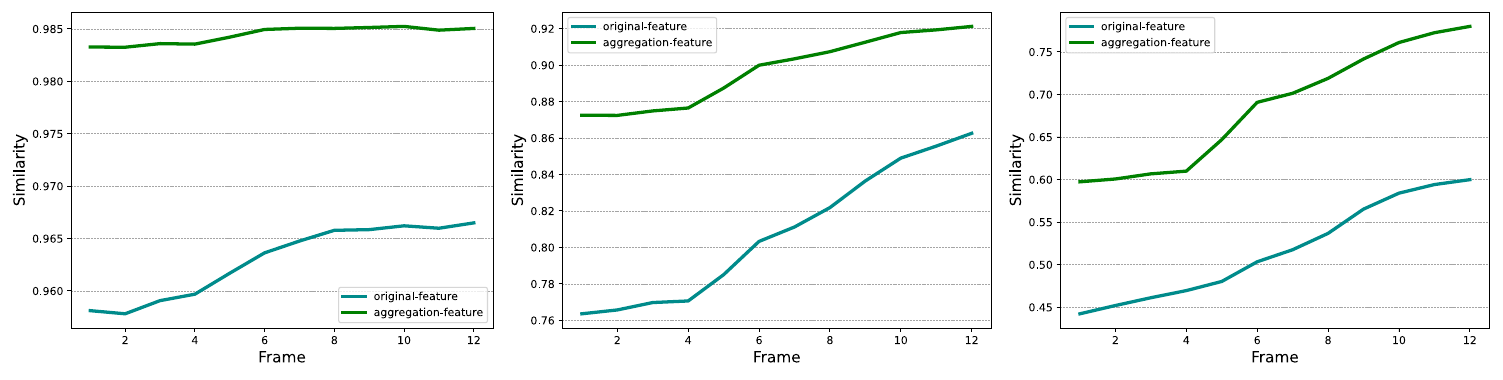}
	\caption{\textbf{The comparison of adjacent frame similarity between original and aggregated features.} The aggregated features from the Feature Cache Module exhibit higher similarity between adjacent frames compared to the original transformer layers, indicating improved temporal coherence.}
	\label{fig_mean_feat_similarity}
\end{figure*}

\subsection{Enhanced Representation for Video Diffusion Model}

Capitalizing on the above discoveries, we introduce RepVideo, a simple and effective framework that leverages enriched intermediate representations to enhance video generation in text-to-video diffusion models. Our approach draws inspiration from recent advancements in diffusion models that incorporate multiple text encoders, such as FLUX~\cite{flux} and MovieGen~\cite{polyak2024movie}. These methods enhance the model's ability to interpret text prompts by employing multiple encoders to capture different layers of information, such as semantic-level and character-level understanding, thereby improving the alignment between generated content and textual descriptions. RepVideo builds upon this idea, aiming to create richer video representations to ultimately improving temporal consistency and semantic alignment in the generated videos.

\begin{figure}[t]
	\centering
	\includegraphics[width=0.4\textwidth]{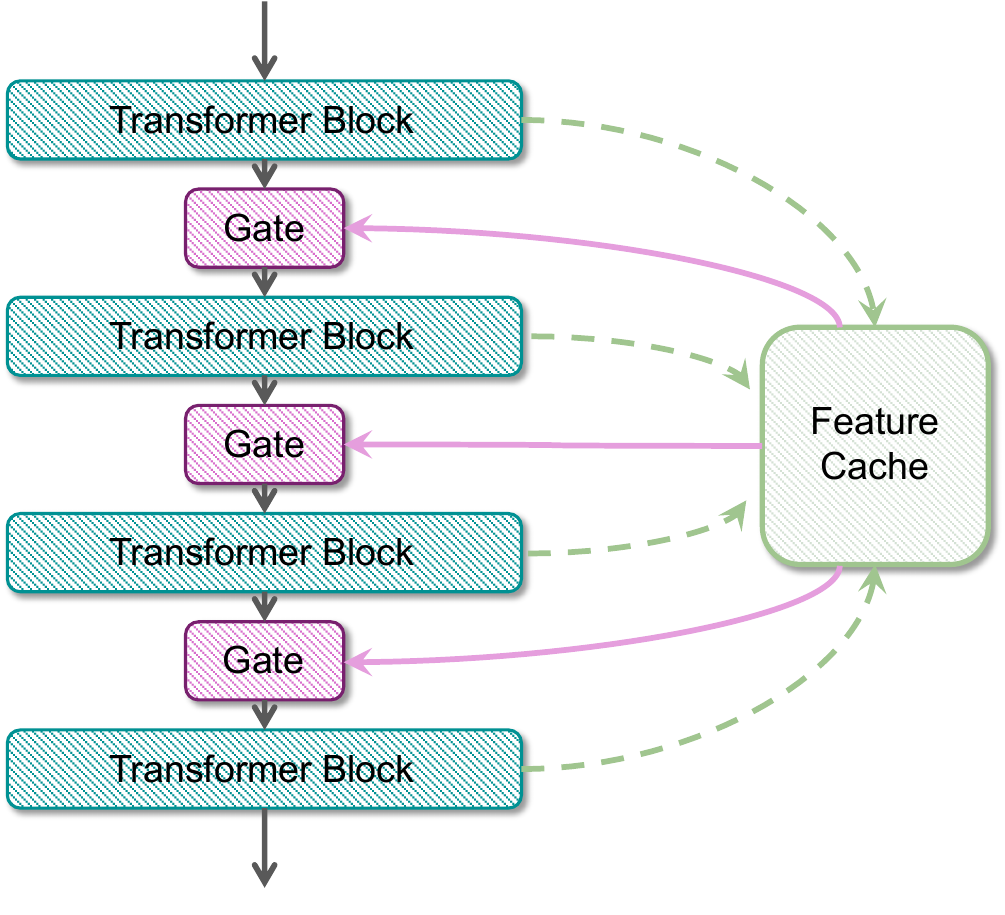}
	\caption{\textbf{The architecture of the enhanced cross-layer
representation framework.}}
	\label{fig_RepVideo_framework}
\end{figure}

To achieve this, we explore leveraging the rich features inherent in diffusion transformers to enhance the semantic consistency and quality of the generated videos. This approach eliminates the need to introduce additional networks, as is done with text encoders, thereby maintaining the model's simplicity and computational efficiency. Firstly, we introduce a Feature Cache Module within the transformer, as illustrated in Fig~\ref{fig_RepVideo_framework}. This module allows each transformer layer to store its output token sequences in the cache,  enabling the Feature Cache Module to aggregate features from multiple adjacent transformer layers. More specifically, we store the output token sequence from the $l$-th transformer layer as follows:
\begin{align}
    \label{eqn_flow_appending}
    \boldsymbol{F}_{orig}^l \to \boldsymbol{\mathcal{M}}_{cache},
\end{align}
where $\boldsymbol{F}_{orig}^l$ represents the output token sequence from the $l$-th layer. Thus, the Feature Cache Module contains the output from multiple layers, aggregating features across different transformer layers. Then, we compute the mean of the features stored in the Feature Cache Module as follows:
\begin{align}
    \label{eqn_flow_appending}
    \boldsymbol{F}_{mean} = \frac{1}{m}\sum_{i=1}^m \boldsymbol{F}_{cache}^i,
\end{align}
where $m$ represents the number of token sequences in the feature cache module. $\boldsymbol{F}_{cache}^i$ is the $i$-th token sequences stored in the cache, \ie $\boldsymbol{F}_{cache}^i \in \boldsymbol{\mathcal{M}}_{cache}$.

We find that by merging these features, a more semantically enriched representation can be achieved. In the Fig~\ref{fig_original_mean_feat}, we present a comparison between the original feature maps produced by a standard transformer layer and those obtained after feature aggregation in the Feature Cache Module.
It can be observed that the aggregated features capture more comprehensive semantic information and exhibit clearer structural information.
Additionally, Fig.~\ref{fig_mean_feat_similarity} illustrates the similarity between adjacent frame features for both sets of representations. The analysis reveals that the aggregated features from the Feature Cache Module exhibit higher similarity between adjacent frames compared to those from the original transformer layers. This suggests that the integration of multi-layer features not only enriches semantic information but also maintains stronger temporal consistency throughout the frames, which is evident in the reduced variability and improved alignment between adjacent frames. 

To use this enriched representation to enhance the original transformer features, we combine the aggregated features with the original input through a gating mechanism:
\begin{align}
    \label{eqn_flow_appending}
    \boldsymbol{F}_{enh}^l = g^l \cdot \boldsymbol{F}_{orig}^l + (1 - g^l) \boldsymbol{F}_{mean}^{l},
\end{align}
where $ \boldsymbol{F}_{mean}^{l}$ represents the aggregated features produced from the feature cache module at $l$-layer. Note that the feature cache module stores a different number of token sequences for each layer, with the number increasing as the layer depth increases. The gating factor $g_l$ is a learnable parameter within the $l$-th layer. This allows the model to dynamically adjust the influence of the enriched representations, effectively balancing the semantic enhancement provided by the cache while preserving necessary layer-specific details from the original feature map.

By using this enriched input, RepVideo enhances the ability of the diffusion model to maintain alignment between the textual input and the generated video, leading to smoother transitions, greater semantic coherence, and an overall improvement in video quality.

\subsection{Training}

To evaluate the effectiveness of the enhanced representation introduced in RepVideo, we implemented our model based on CogVideoX-2B~\cite{yang2024cogvideox}. This baseline was selected due to its robust architecture and proven performance in text-to-video generation tasks, providing a solid foundation for comparison. The training process was carefully designed to ensure fairness and consistency between the baseline and our proposed method.\\[1ex]
\noindent \textbf{Data Preparation.} 
Since CogVideoX-2B~\cite{yang2024cogvideox} is pretrained on a large-scale corpus of videos, we constructed a curated internal dataset tailored for fine-tuning the model. The dataset, sourced from high-quality platforms 
underwent a rigorous preprocessing pipeline to ensure diversity and quality. First, long videos were segmented into shorter, manageable clips to emphasize focused events or actions. Related clips were then linked together to form coherent event sequences, preserving narrative consistency across frames. To refine the dataset further, static video filtering was applied to exclude clips lacking significant motion, ensuring an emphasis on dynamic content.

Additionally, aesthetic scoring was conducted to evaluate visual quality based on predefined criteria, prioritizing high-quality and semantically rich video inputs. Dynamic estimation was employed to analyze movement patterns and overall clip dynamics, enhancing the dataset's relevance for motion-centric video generation. A watermark classifier was applied to detect and annotate videos containing visible watermarks, maintaining the integrity and usability of the training data. This comprehensive data preparation process resulted in a high-quality dataset of 1 million annotated video clips with detailed captions, covering a diverse range of categories.\\[1ex]
\noindent \textbf{Training Setup.} 
Both the baseline CogVideoX-2B~\cite{yang2024cogvideox} and the enhanced RepVideo model were fine-tuned under identical training conditions to ensure fair comparisons. The models were trained for 50,000 steps with a batch size of 32. The AdamW optimizer was employed, using a learning rate of $1 \times 10^{-5}$, which balanced efficient convergence with model stability. Training was conducted on 32 NVIDIA H100 GPUs, leveraging the computational resources required for large-scale video generation tasks. A separate validation set, curated from the prepared dataset, was utilized to periodically assess model performance during training and mitigate the risk of overfitting. These measures ensured that the training process was rigorous and reproducible.\\[1ex]
\noindent \textbf{Enhanced Representation Implementation.} 
The core innovation of RepVideo lies in its ability to aggregate intermediate transformer outputs to create stable and semantically enriched representations, improving both spatial fidelity and temporal consistency. This was achieved through the introduction of a Feature Cache Module, which aggregates features across every $m$ transformer layers. By aggregating features from neighboring layers, the module effectively captures semantic details while mitigating inconsistencies across frames. Empirical evaluations demonstrated that setting $m = 6$ provided the optimal balance between computational efficiency and performance improvement. Unless otherwise specified, $m = 6$ is used throughout the experiments.

To integrate these enriched representations into the transformer layers, a gating mechanism was introduced. The gating mechanism dynamically combines the aggregated features with the original transformer outputs, using a learnable parameter to control their relative influence. 

The proposed aggregation and gating mechanisms are lightweight, introducing minimal additional parameters and computational overhead. During training, these mechanisms demonstrated significant improvements in spatial consistency, temporal coherence, and semantic alignment, resulting in smoother and more visually compelling video outputs.

%% file: sec/4_experiments.tex
\begin{table*}[t]
\small
\centering
    \caption{Comparison with previous methods on VBench~\cite{vbench} .}
    \begin{tabular}{lccccc}
    \toprule
     Method  & Total Score & Motion Smoothness & Object Class &Multiple Objects & Spatial Relationship  \\
    \midrule
      LaVie~\cite{wang2023lavie} & 77.08\% & 96.38\% & 91.82\% & 33.32\% & 34.09\%  \\
      VideoCrafter-2.0~\cite{chen2024videocrafter2} & 80.44\% & 97.73\% & 92.55\% & 40.66\% & 35.86\%  \\
      OpenSoraPlan-v1.1 & 78.00\% & 98.28\% & 76.30\% & 40.35\% & 53.11\%  \\
      OpenSora-1.2~\cite{lin2024opensoraplanopensourcelarge} & 79.76\% & 98.50\% & 82.22\% & 51.83\% & 68.56\%  \\
      Gen-3~\cite{Gen-3} & 82.32\% & 99.23\% & 87.81\% & 53.64\% & 65.09\% \\
      Gen-2~\cite{Gen-2} & 80.58\% & 99.58\% & 90.92\% & 55.47\% & 66.91\% \\
      Pika-1.0~\cite{Pika} & 80.69\% & 99.50\% & 88.72\% & 43.08\% & 61.03\% \\
      CogVideoX-2B~\cite{yang2024cogvideox} & 80.91\% & 97.73\% & 83.37\% & 62.63\% & 69.90\%  \\
      CogVideoX-5B~\cite{yang2024cogvideox}  & 81.61\% & 96.92\% & 85.23\% & 62.11\% & 66.35\%  \\
      Vchitect-2.0~\cite{Vchitect} & 81.57\% & 97.76\% & 87.81\% & 69.35\% & 54.64\% \\
     \midrule
     \rowcolor{gray!20} \textbf{RepVideo} (Ours)  & 81.94\% & 98.13\% & 87.83\% & 71.18\% & 74.74\% \\
    \bottomrule
    \end{tabular}
\label{tab:mini}
\end{table*}
\begin{table*}[t]
\small
\centering
    \caption{User study between CogVideoX-2B~\cite{yang2024cogvideox} and RepVideo.}
    \begin{tabular}{lcccc}
    \toprule
     Method  & Overall Score & Spatial Appearance  & Temporal Consistency  & Video-text alignment  \\
    \midrule
      CogVideoX-2B~\cite{yang2024cogvideox} & 25.54\% & 25.00\% & 25.93\% & 23.38\%   \\
      RepVideo & 75.46\% & 75.00\% & 74.07\% & 76.62\%  \\
    \bottomrule
    \end{tabular}
\label{tab:user_study}
\end{table*}
\section{Experiments}
To thoroughly evaluate the effectiveness of our proposed framework, we conducted a comprehensive set of experiments covering both quantitative and qualitative analyses. These experiments were designed to assess the performance of RepVideo across a variety of metrics, including spatial fidelity, temporal consistency, and semantic alignment with textual prompts. In particular, we benchmark our model against state-of-the-art methods in automated metrics and human preference evaluations, providing a comprehensive view of its advantages.

\begin{figure*}[!h]
	\centering
	\includegraphics[width=0.99\textwidth]{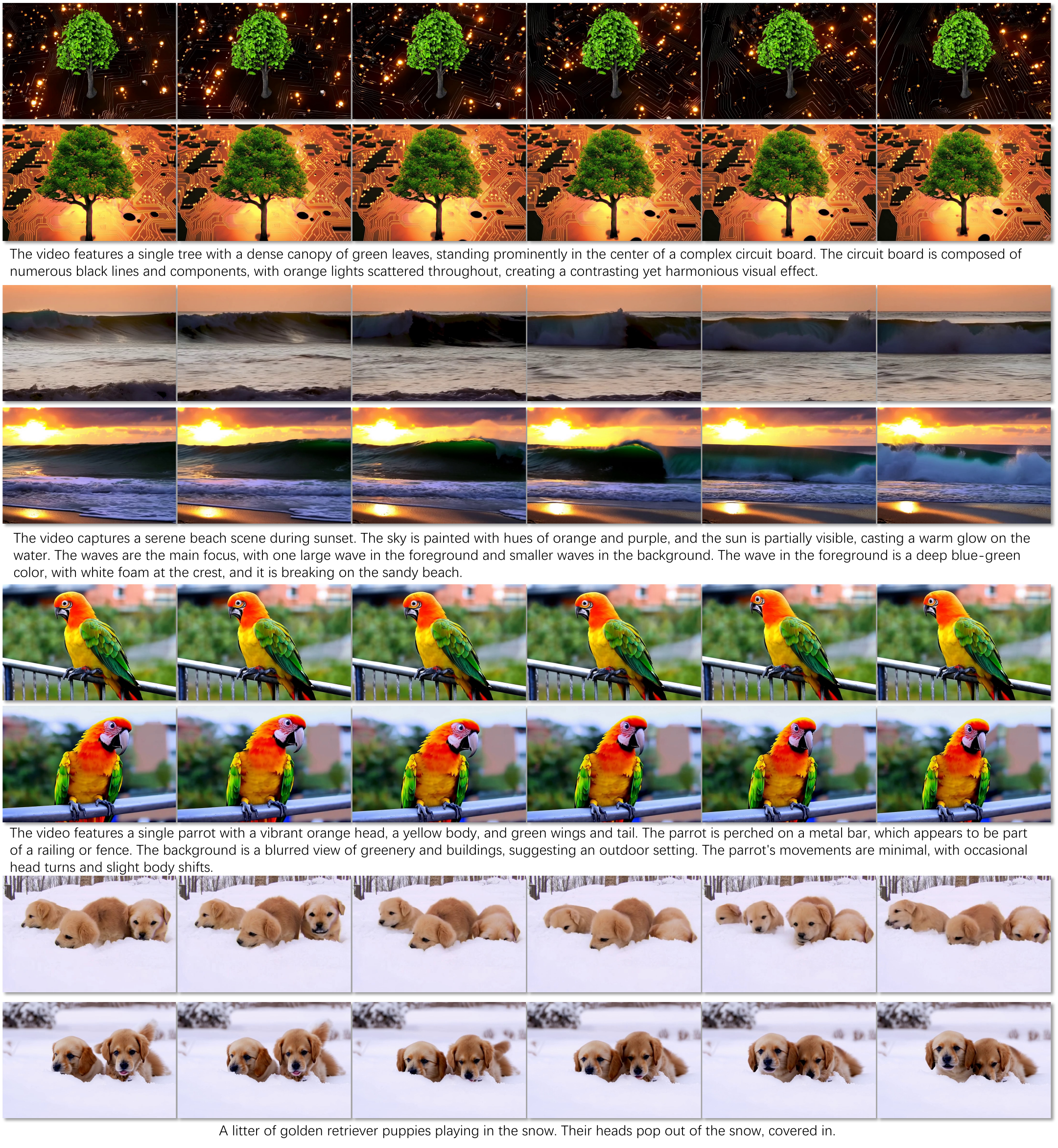}
	\caption{\textbf{The qualitative comparison between our method and the baseline CogVideoX-2B~\cite{yang2024cogvideox}.} The first row shows results from the baseline CogVideoX-2B~\cite{yang2024cogvideox}, while the second row presents the results generated by RepVideo, demonstrating significant improvements in quality and coherence.}
	\label{fig_rep_cog_video}
\end{figure*}

The experiments are organized into three parts: \textbf{1).} Automated Evaluation: Quantitative metrics derived from VBench~\cite{vbench} are used to objectively compare RepVideo’s performance with existing models. \textbf{2).} Human Evaluation: Human raters assess the generated videos based on alignment with prompts, temporal smoothness, and frame-wise quality, offering a complementary perspective. \textbf{3).} Ablation Study: A detailed investigation into the contributions of the design of RepVideo, analyzing how spatial and temporal consistency are improved. The following sections delve into these evaluations, highlighting both the strengths and potential areas of improvement for our approach.
\subsection{Automated Evaluation}
\noindent \textbf{Quantitative Evaluation}. To evaluate the performance of our model, we adopt all metrics provided in VBench~\cite{vbench, huang2024vbenchplusplus}. As shown in Table~\ref{tab:mini}, we report the \textit{Total Score} alongside several representative metrics. Specifically, \textit{Motion Smoothness} evaluates the temporal stability of generated videos, while \textit{Object Class} and \textit{Multiple Objects} measure the ability to generate diverse and well-defined visual elements. Additionally, \textit{Spatial Relationship} assesses the coherence in object positioning and interactions.

Compared to the baseline model CogVideoX-2B, our model, RepVideo-2B, achieves superior results with a higher \textit{Total Score}. Notably, our model improves by \textbf{0.4\%} in \textit{Motion Smoothness} and \textbf{4.46\%} in \textit{Object Class}, highlighting its ability to maintain temporal consistency and generate fine-grained object details. Moreover, RepVideo achieves significant improvements in \textit{Spatial Relationship} (+\textbf{4.84\%}) and \textit{Multiple Objects} (+\textbf{8.55\%}), demonstrating enhanced spatial coherence and strong ability to handle complex object interactions.\\[1ex]
\noindent \textbf{Qualitative Evaluation}.
Figure~\ref{fig_rep_cog_video} provides a qualitative comparison between our method and the baseline, CogVideoX-2B~\cite{yang2024cogvideox}, demonstrating the significant improvements achieved by our model. The results generated by our method are presented in the second column, while the baseline’s outputs are displayed in the first column. It is obvious that our approach produces more visually consistent and semantically accurate videos, capturing both the spatial and temporal relationships described in the provided prompts. For example, with the prompt “A litter of golden retriever puppies playing in the snow. Their heads pop out of the snow, covered in,” our model maintains temporal coherence, ensuring consistent appearance and motion of the puppies across frames. In contrast, the baseline struggles with stability, resulting in artifacts and inconsistencies. In the prompt “The video features a serene beach scene during sunset,” our method effectively captures the smooth motion of the sunset and spatial accuracy, while the baseline fails to understand the sunset. Similarly, for “A single tree with a dense canopy of green leaves, standing prominently in the center of a complex circuit board,” our method preserves both the intricate details and spatial alignment across frames. The baseline, however, exhibits noticeable jitter and spatial inconsistencies. Finally, with “A single parrot with a vibrant orange head, a yellow body, and green wings and tail,” our model ensures vivid colors and smooth transitions, maintaining temporal stability. The baseline fails to generate coherent movements, leading to visual distortions. These qualitative results underscore the strength of our proposed RepVideo framework in generating high-quality videos with enhanced temporal coherence and spatial fidelity.

\subsection{Human Evaluation}
In addition to the automated evaluation, we conducted a comprehensive human evaluation to assess the performance of our model in comparison to state-of-the-art methods. Human evaluators were presented with pairs of videos generated by different models, each conditioned on the same text prompt. Evaluators were tasked with selecting their preferred video based on three key criteria, evaluated independently: video-text alignment, temporal consistency, and spatial appearance. These criteria were derived from the dimension design in VBench~\cite{huang2024vbenchplusplus}, ensuring a standardized and rigorous evaluation process.

The evaluation included two models: our proposed RepVideo and CogVideoX-2B~\cite{yang2024cogvideox}. For each criterion, evaluators conducted 50 pairwise comparisons between videos generated by our model and those produced by the competing method.

As summarized in Tab.~\ref{tab:user_study}, our model achieved an average win ratio exceeding 50\% across all three metrics, demonstrating its superiority in generating videos with higher semantic alignment, smoother temporal transitions, and enhanced visual quality. These results underscore the effectiveness of the enriched representations introduced in our framework, as evidenced by the clear preference expressed by human evaluators.

\begin{figure*}[t]
	\centering
	\includegraphics[width=0.99\textwidth]{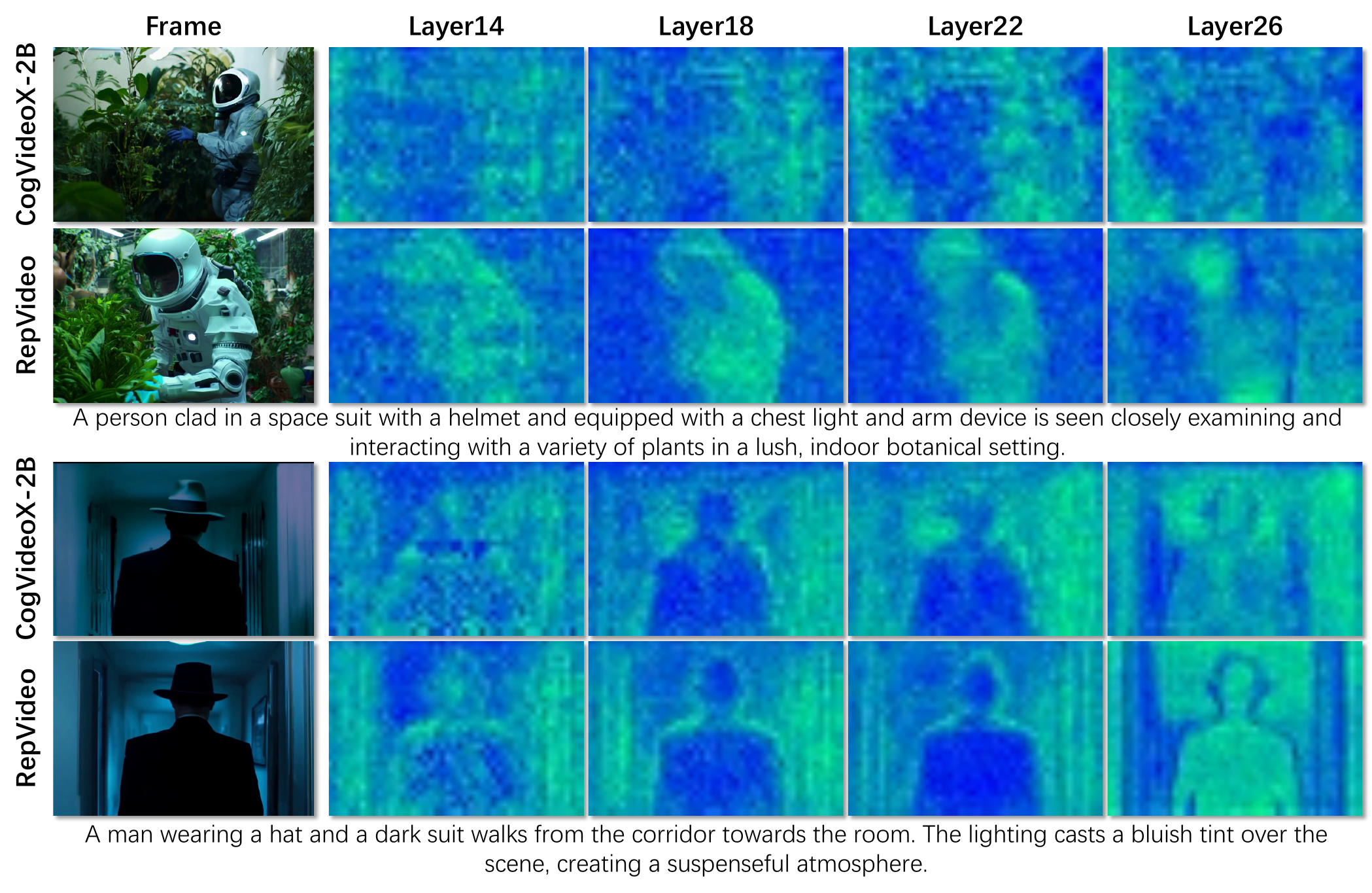}
	\caption{\textbf{The Layer-wise comparison of feature maps between CogVideoX-2B and RepVideo.} The comparison shows that RepVideo consistently captures richer semantic information and maintains more coherent spatial details across layers compared to CogVideoX-2B~\cite{yang2024cogvideox}.}
	\label{fig_feature_maps_cog_rep}
\end{figure*}

\subsection{Ablation Study}

To validate the effectiveness of our proposed method, we conducted two experiments that highlight how RepVideo improves both spatial appearance and temporal consistency. These evaluations combine qualitative visualizations and quantitative metrics to demonstrate the advantages of our approach.

\noindent \textbf{How RepVideo Improves Spatial Appearance.}
Figure~\ref{fig_feature_maps_cog_rep} provides a layer-wise comparison of feature maps between CogVideoX-2B~\cite{yang2024cogvideox} and RepVideo. The results demonstrate that our model consistently captures richer semantic information and maintains more coherent spatial details as the layers deepen. For instance, in the example “A man wearing a hat and a dark suit walks from the corridor towards the room,” the feature maps generated by RepVideo clearly preserve the structure and contours of the man, ensuring that the generated video retains sharp and well-defined spatial attributes. The deeper layers of CogVideoX-2B~\cite{yang2024cogvideox}, in contrast, display feature maps that are blurred and lack focus, failing to capture key semantic elements of the scene.

\begin{figure*}[t]
	\centering
	\includegraphics[width=0.95\textwidth]{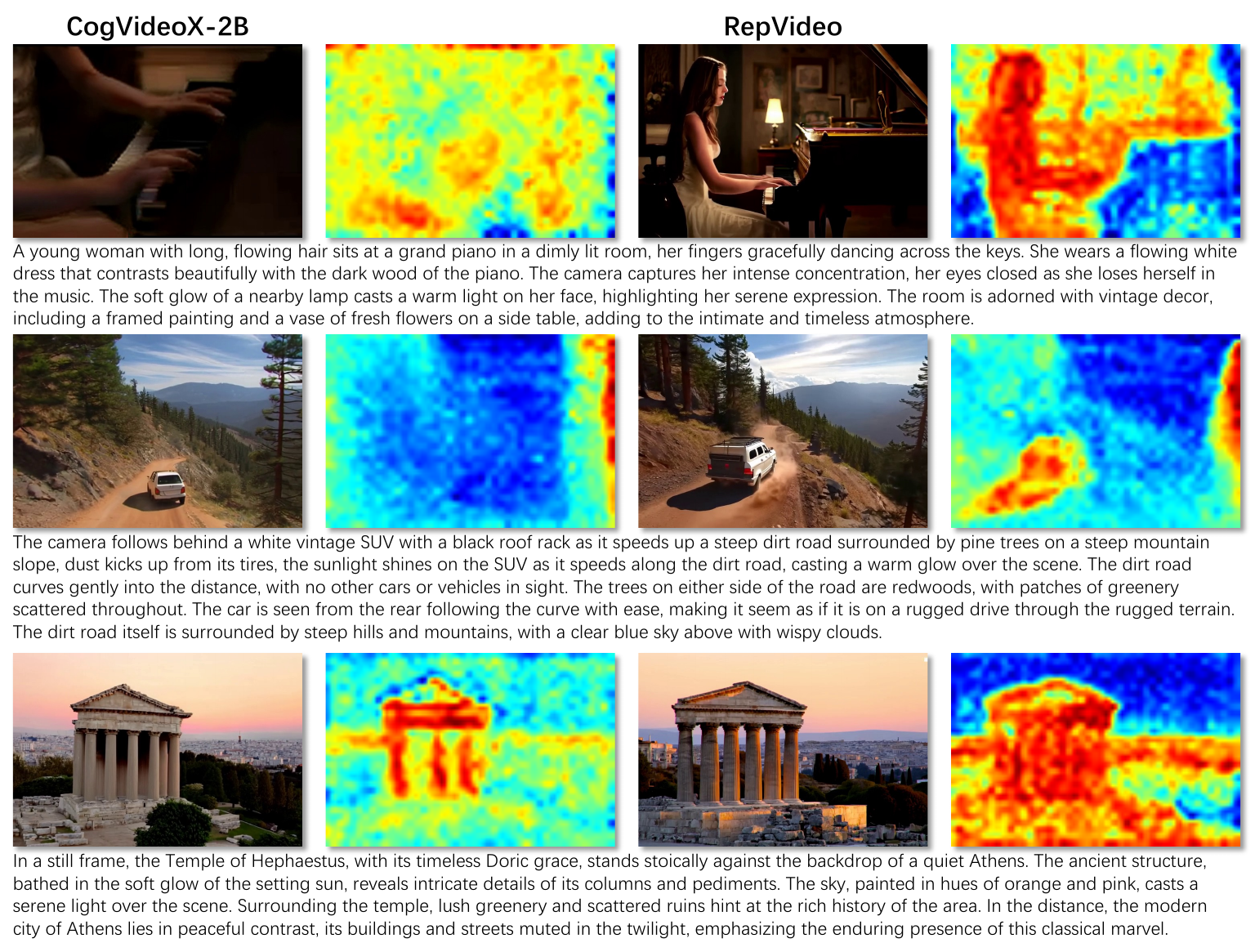}
	\caption{\textbf{The comparison of attention maps between CogVideoX-2B and RepVideo.} The comparison demonstrates that RepVideo could maintain more consistent semantic relationship compared to CogVideoX-2B~\cite{yang2024cogvideox}.}
	\label{fig_comp_attention_maps}
\end{figure*}

Similarly, in the example “A person clad in a space suit is seen interacting with a variety of plants in a lush, indoor botanical setting,” RepVideo’s feature maps demonstrate a superior ability to capture fine-grained details, such as the distinct outlines of the astronaut and the plants. As the layer depth increases, our model retains semantic consistency and spatial integrity, whereas CogVideoX-2B~\cite{yang2024cogvideox} struggles with spatial coherence, often leading to visual artifacts in the generated videos. These results underscore the importance of RepVideo’s enriched representations, which aggregate information across layers to stabilize spatial details and enhance semantic alignment.

\begin{figure*}
	\centering
	\includegraphics[width=0.99\textwidth]{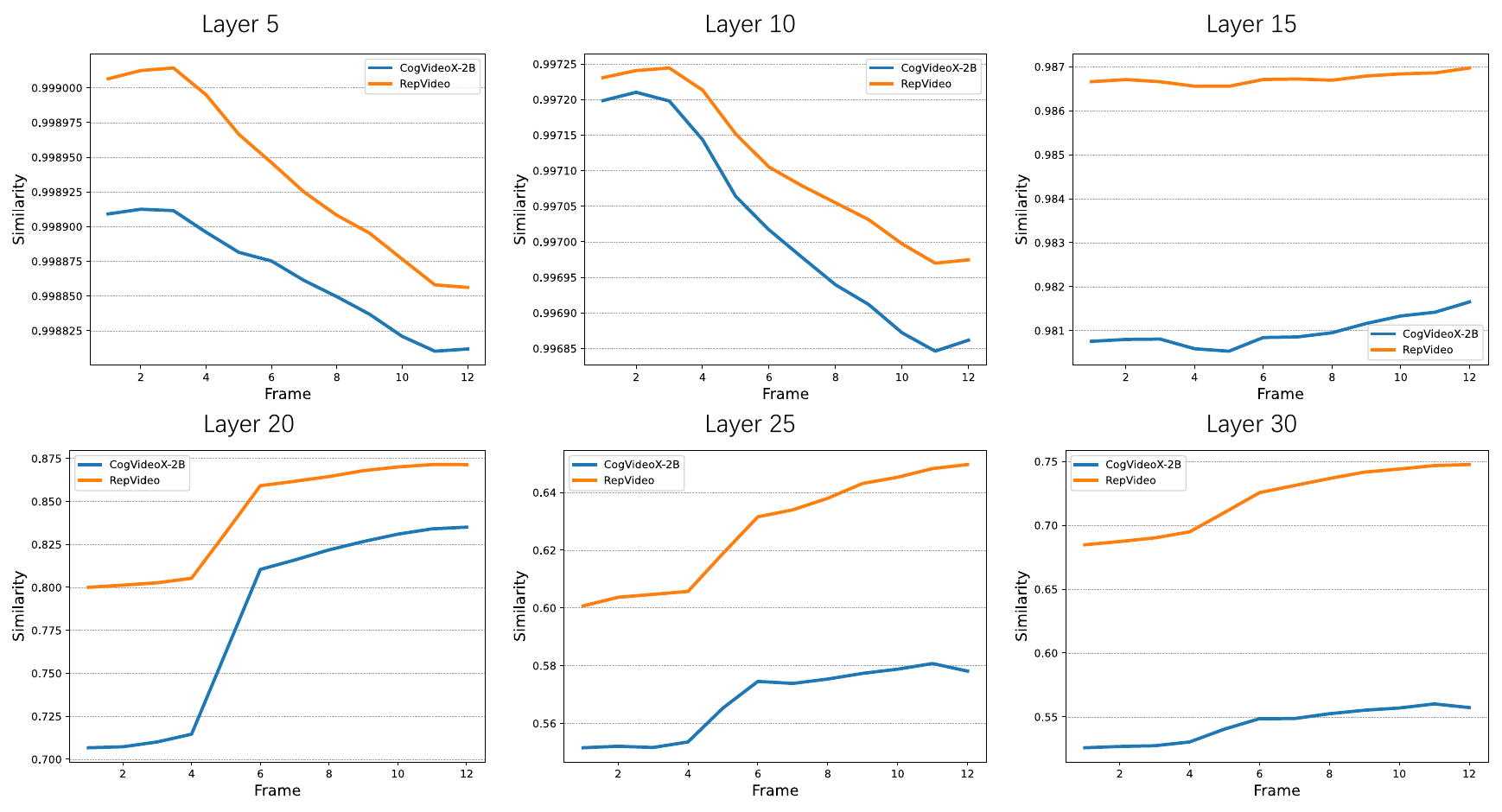}
	\caption{\textbf{The cosine similarity between consecutive frames across layers.}}
	\label{fig_similarity_cog_repv}
\end{figure*}

The improved spatial appearance can be attributed to our feature aggregation mechanism, which leverages intermediate representations from consecutive layers. As shown in Figure~\ref{fig_comp_attention_maps}, the visualization of the attention map demonstrates whether the objects in the frame are activated under different prompts. The attention map of our RepVideo clearly shows the boundary of different subjects mentioned in the prompts compared with CogVideoX, demonstrating that our aggregated features can help to strengthen the corresponding spatial regions. By combining these features, RepVideo reduces variability across layers and ensures that critical spatial information is preserved throughout the video generation process. This mechanism enhances the model’s ability to generate scenes that are visually consistent and aligned with the input prompts.\\[1ex]
\noindent \textbf{How RepVideo Improves Temporal Consistency.}
In addition to improving spatial details, RepVideo significantly enhances temporal consistency, a key factor in generating coherent videos. To quantify this improvement, we calculated the cosine similarity between consecutive frames, as shown in Figure~\ref{fig_similarity_cog_repv}. The x-axis represents the layer depth, while the y-axis indicates the cosine similarity between the $n^{\text{th}}$ frame and the $(n+1)^{\text{th}}$ frame. Higher similarity values suggest better temporal stability, with smaller variations between consecutive frames.

The results demonstrate that RepVideo achieves consistently higher cosine similarity scores across all layers compared to CogVideoX-2B~\cite{yang2024cogvideox}. For example, at deeper layers such as Layer 25 and Layer 30, RepVideo maintains a strong frame-to-frame similarity, ensuring that the generated videos are free from temporal artifacts like flickering or jittering. In contrast, CogVideoX-2B~\cite{yang2024cogvideox} shows a sharp decline in similarity as the layer depth increases, indicating weaker temporal consistency and a higher likelihood of motion discontinuities.

The enhanced temporal consistency achieved by RepVideo is due to its innovative feature cache module. By aggregating features from multiple adjacent layers, the module stabilizes intermediate representations and minimizes temporal variability. This approach ensures that each frame is not only semantically aligned with the input prompt but also consistent with preceding and succeeding frames. As a result, RepVideo produces videos with smooth transitions and coherent motion, even in complex scenarios involving dynamic objects or environments.

%% file: sec/5_conclusion.tex
\section{Discussion}

While \textbf{RepVideo} achieves notable advancements over existing methods, there are areas where our approach could be improved. A key limitation comes from our reliance on pre-trained models, such as CogVideoX-2B~\cite{yang2024cogvideox}. These models, while robust, carry inherent biases and constraints from their original training datasets, which can limit the diversity and adaptability of the generated videos, particularly in scenarios that fall outside the training data distribution.

Another area for improvement is the computational cost associated with the feature aggregation mechanism. Although the approach is relatively lightweight compared to adding new parameters, it may still present challenges for real-time applications or deployment in environments with limited resources. Finding ways to streamline this mechanism without compromising its effectiveness remains an important direction for future work.

Moreover, the method could perform better in generating human-centric content and understanding complex spatial relationships. As highlighted in our experiments, \textbf{RepVideo} occasionally struggles to maintain consistency in videos that require precise modeling of human actions or intricate object arrangements.

Future research could focus on developing real-time feature aggregation mechanism. Additionally, integrating our method with different text-to-video pretrained-model could be beneficial to the whole community. Addressing these limitations could further expand the potential of \textbf{RepVideo}, enabling it to set new benchmarks in text-to-video generation tasks.

\section{Conclusion}

This paper presented \textbf{RepVideo}, a framework designed to improve text-to-video diffusion models by addressing challenges in spatial appearance and temporal consistency. Our approach leverages enriched intermediate representations, combining feature aggregation and gating mechanisms to enhance semantic stability across frames and maintain fine-grained spatial details. Built on the solid foundation of CogVideoX-2B~\cite{yang2024cogvideox}, \textbf{RepVideo} demonstrates substantial improvements in generating visually consistent and semantically aligned videos.

Through extensive experiments, we validated the effectiveness of our framework. \textbf{RepVideo} outperforms state-of-the-art models across automated benchmarks and human evaluations, showing its ability to maintain temporal coherence in dynamic scenarios and to generate detailed, accurate spatial representations. These contributions represent a meaningful advancement in text-to-video generation, providing a strong platform for future innovation and application in this evolving field.